\pdfoutput=1
\documentclass[letterpaper, 10 pt, conference]{ieeeconf}  

\let\labelindent\relax

\IEEEoverridecommandlockouts                              

\title{\LARGE \bf
QuadWBG: Generalizable Quadrupedal Whole-Body Grasping}

\author{
Jilong Wang$^{1,2*}$,
Javokhirbek Rajabov$^{1,2*}$,
Chaoyi Xu$^{2,4}$,
Yiming Zheng$^{2, 3}$,
He Wang$^{1,2,4\dagger}$
\thanks{*:joint first authors, $\dagger$: corresponding author, email: hewang@pku.edu.cn}
\thanks{
$^{1}$CFCS, School of Computer Science, Peking University.}
\thanks{
$^{2}$Galbot \quad\quad
$^{3}$University of Toronto }
\thanks{$^{4}$Beijing Academy of Artificial Intelligence (BAAI)}
}

\usepackage{graphicx}
\usepackage{subcaption}
\usepackage[table]{xcolor}
\usepackage{array}
\usepackage{booktabs}
\usepackage{tabularx}
\usepackage{float}
\usepackage{times}
\usepackage{cite}
\usepackage{url}
\usepackage{mathtools}
\usepackage{amsmath}

\usepackage{amssymb}
\usepackage{amsfonts}
\usepackage{amsopn}
\usepackage{graphicx} 
\usepackage{textcomp}
\usepackage{xfrac}
\usepackage{bbm}
\usepackage{overpic}
\usepackage{wrapfig}
\usepackage[ruled,vlined,linesnumbered]{algorithm2e}
\SetAlFnt{\small}
\SetAlCapFnt{\small}
\SetAlCapNameFnt{\small}
\SetAlCapHSkip{0pt}

\usepackage{multirow}

\usepackage{booktabs}
\usepackage{footnote}
\usepackage{paralist}
\usepackage{enumitem}
\usepackage{autobreak}
\usepackage{hyperref}
\usepackage{cleveref}

\usepackage{color}
\definecolor{green}{rgb}{0, 0.4, 0}
\definecolor{orange}{rgb}{0.8, 0.6, 0.2}
\definecolor{red}{rgb}{1.0, 0.0, 0.0}
\definecolor{teal}{rgb}{0.0, 0.4, 0.4}
\definecolor{purple}{rgb}{0.65,0,0.65}
\definecolor{saffron}{rgb}{0.95,0.75,0.2}
\definecolor{turquoise}{rgb}{0.0,0.5,0.5}
\definecolor{brown}{rgb}{0.5, 0.16, 0.16}

\usepackage{overpic}
\usepackage{currfile} 

\newlength\savedwidth

\definecolor{lightgray}{rgb}{0.6, 0.6, 0.6}

\newcommand{\addcite}[1]{{\textcolor{red}{[cite]}}}

\definecolor{revisedcolor}{RGB}{100,0,200}

\usepackage[normalem]{ulem}

\newcommand{\hidecomment}[1]{}

\usepackage{xspace}

\begin{document}

\maketitle
\thispagestyle{empty}
\pagestyle{empty}


\begin{abstract}
Legged robots with advanced manipulation capabilities have the potential to significantly improve household duties and urban maintenance. Despite considerable progress in developing robust locomotion and precise manipulation methods, seamlessly integrating these into cohesive whole-body control for real-world applications remains challenging. In this paper, we present a  modular framework for robust and generalizable whole-body loco-manipulation controller based on a single arm-mounted camera. By using reinforcement learning (RL), we enable a robust low-level policy for command execution over 5 dimensions (5D) and a grasp-aware high-level policy guided by a novel metric, Generalized Oriented Reachability Map (GORM). The proposed system achieves state-of-the-art one-time grasping accuracy of 89\% in real world, including challenging tasks such as grasping transparent objects. 
Through extensive simulations and real-world experiments, we demonstrate that our system can effectively manage a large workspace, from floor level to above body height, and perform diverse whole-body loco-manipulation tasks. See our robot at work: \url{quadwbg.github.io}.

\end{abstract}
\section{INTRODUCTION}

Quadrupedal loco-manipulation, which integrates legged locomotion with robotic arm manipulation, has emerged as a key research area due to its broad potential applications, including household assistance, urban maintenance, disaster relief, and autonomous field operations \cite{lindqvist2022multimodality, arm2023scientific, stachowiak2021procedures, hoeller2024anymal}. Recent advancements in reinforcement learning (RL) have enabled the development of end-to-end policies for whole-body locomotion and manipulation \cite{fu2023deep, liu2024visual, portela2024learning, pan2024roboduet, ha2024umi}, allowing robots to perform tasks that require seamless coordination of movement and object interaction. While end-to-end RL has substantially improved locomotion skills \cite{margolis2024rapid, ji2022concurrent, fu2021minimizing, hwangbo2019learning, he2024agile, margolis2021learning, li2023robust, yang2020multi, ma2023learning, zhuang2023robot, cheng2024extreme, liao2023walking, miki2022learning}, loco-manipulation remains highly challenging due to the increased action dimensionality and complex physical interactions involved. These challenges often result in loco-manipulation policies with mediocre accuracy and limited generalizability \cite{fu2023deep, portela2024learning}, especially when grasping objects of different shapes, sizes, and materials, thereby restricting their effectiveness in real-world applications.

\begin{figure}[t]
\centering
\includegraphics[bb=0 0 3580 2000, width=0.48\textwidth]{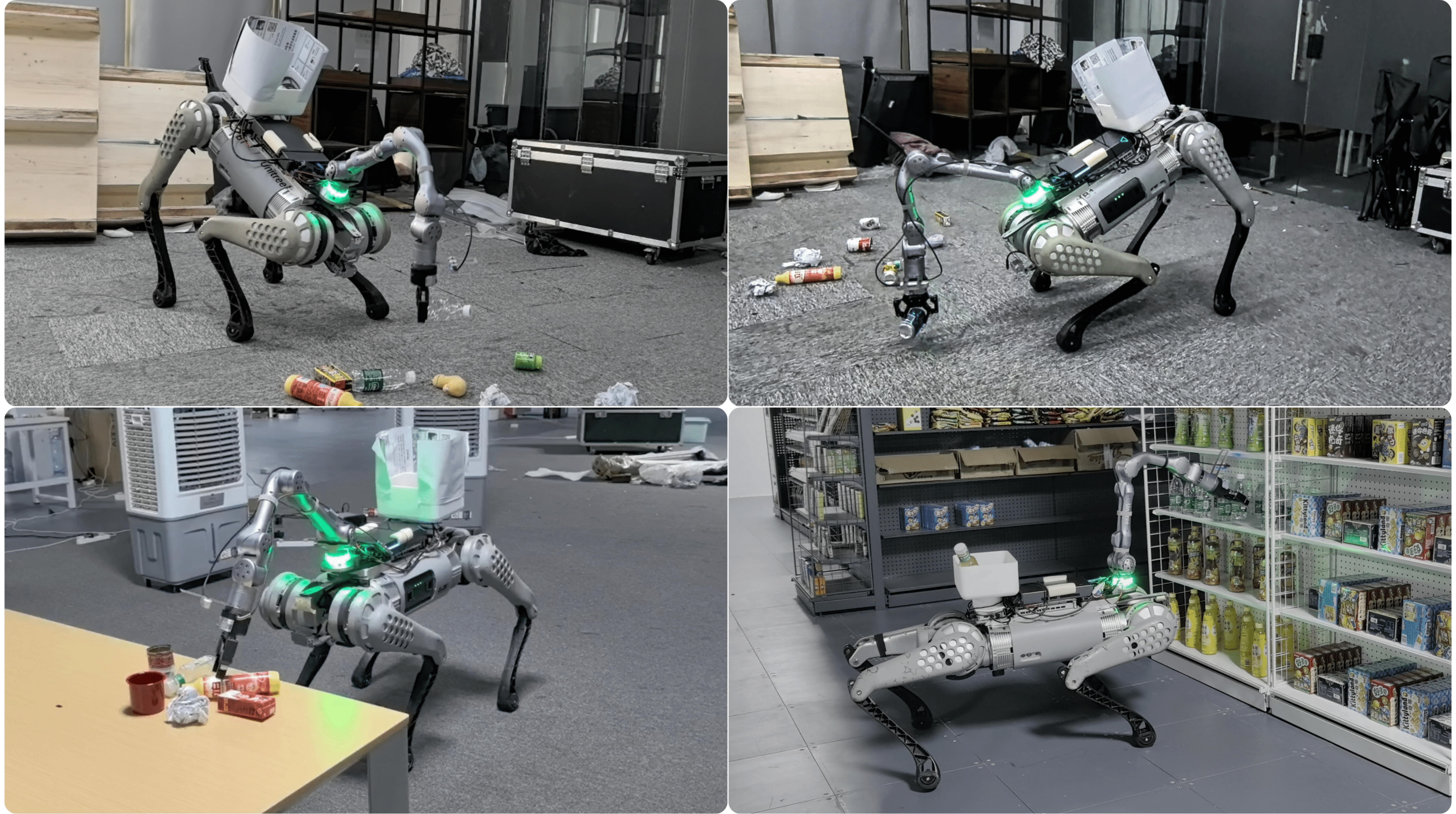}

\caption{ Real-world examples of whole-body grasping objects from various locations. \textbf{Top row}: Grasping transparent and specular objects in cluttered environments. \textbf{Bottom row}: Grasping of objects at various heights.}
\label{fig:grasping_fusion}
\end{figure}

To enhance both the performance and generalizability of whole-body grasping systems, we draw inspiration from the success of various grasp detection techniques \cite{wang2021graspness, fang2020graspnet, shi2024asgrasp, sundermeyer2021contact, dai2022domain, fang2023anygrasp, kuang2024ram}. These methods demonstrate robust performance in detecting grasp poses for diverse, unseen objects in cluttered environments, including challenging materials like transparent or specular surfaces. By integrating grasp pose detection with motion planning, these methods consistently demonstrate impressive accuracy, achieving whole-body grasping success rates of approximately 90\% across diverse object instances, configurations, and environments, as illustrated in \ref{fig:grasping_fusion}.

This inspires us to take the best of both worlds via integrating legged locomotion with grasp detection to achieve high-performance and highly generalizable loco-manipulation. However, this combination is highly nontrivial. Directly applying grasp detection results for arm motion planning in legged robots is insufficient, as it ignores the coordination required between the robot's body and arm movements.

\begin{figure*}[ht]
\centering
\includegraphics[bb=0 0 3285 2730, trim=0 10 0 10, clip, width=0.85\textwidth]{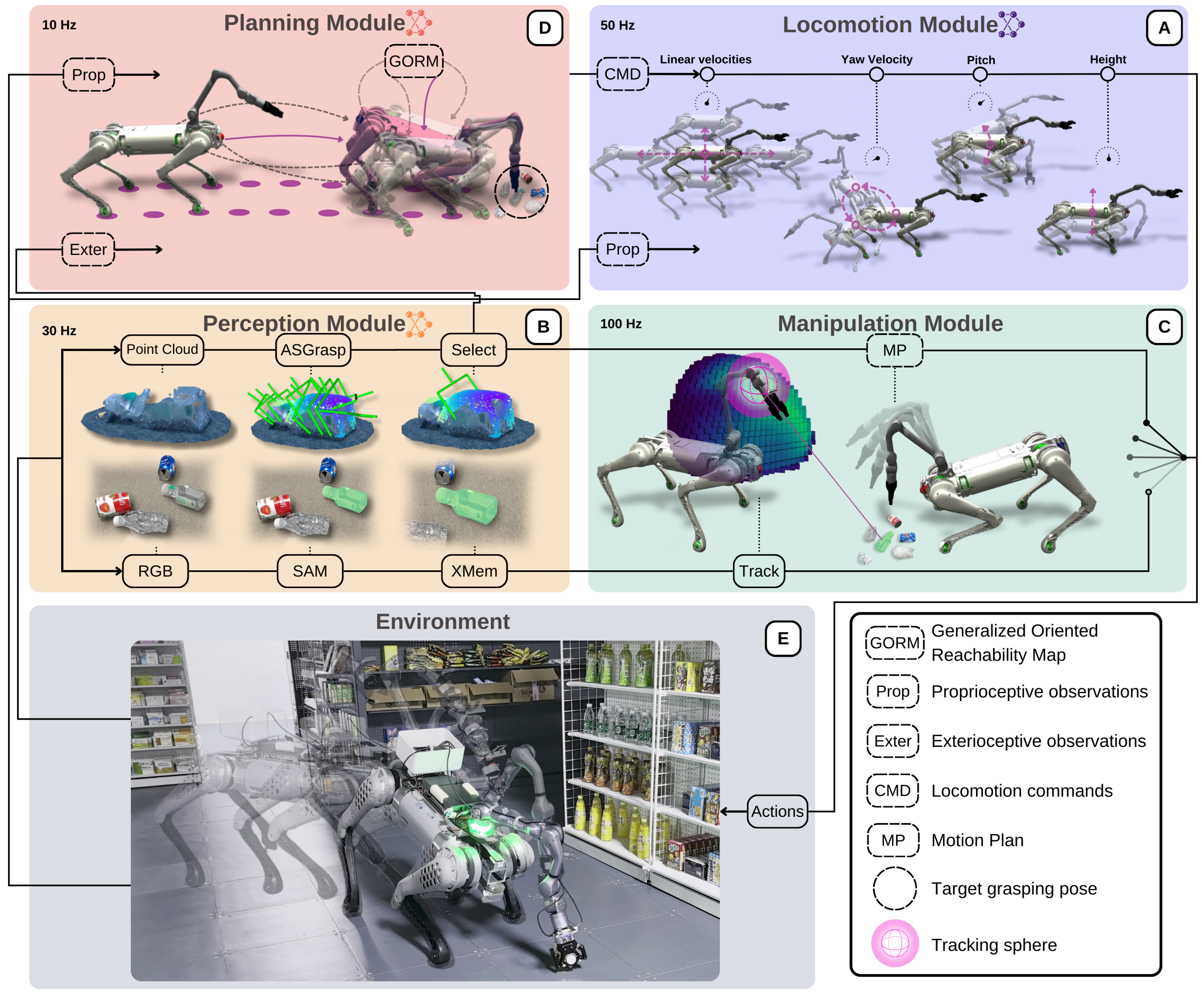}
\caption{Pipeline of our system. 
First, we train a teacher-student 5D low-level policy in simulation (A). The perception module (B) then continuously tracks the object, and generates grasping pose guiding manipulation module (C). This pose is also utilized by the planning module (D) to command the locomotion policy.}
\label{fig:pipeline}
\end{figure*}

To address these challenges, we introduce \textbf{QuadWBG} (Generalizable \textbf{Quad}rupedal \textbf{W}hole-\textbf{B}ody \textbf{G}rasping), a modular system consisting of four key modules: perception, planning, locomotion, and manipulation (see Figure \ref{fig:pipeline}). The perception module integrates object segmentation and grasp detection, handling object tracking and grasp pose prediction. The planning and locomotion modules function as high-level and low-level controllers, respectively, guiding the robot to approach the grasp pose. Finally, the manipulation module leverages motion planning to move the arm and execute the grasp while maintaining the body stationary. 

At the core of this system is a key innovation: the \textbf{Generalized Oriented Reachability Map (GORM)}. GORM acts as a metric for evaluating the reachability of the base pose relative to the target pose across six degrees of freedom. It efficiently guides the planning module during training by calculating the optimal base pose for grasping tasks. GORM also captures the robot’s overall reachability from various positions and orientations, enabling the policy to select base poses that optimize both arm reachability and robot stability. This ensures precise grasping while maintaining balance and locomotion stability.

Our framework offers additional advantages. By incorporating highly robust grasp detection modules trained on large-scale real and synthetic data, we eliminate the sim-to-real gap commonly encountered in end-to-end visual policy learning for manipulation. Furthermore, our framework provides clearer insights into system performance and allows for targeted optimization of each module.

Through extensive simulations and real-world experiments, we demonstrate that our system achieves state-of-the-art performance in both grasping accuracy and handling a wide range of objects. The system consistently delivers robust whole-body control across a variety of tasks. Notably, it achieves a remarkable one-time grasping accuracy of 89\% in real-world scenarios, even excelling in challenging tasks such as grasping transparent objects. These results underscore the system's high precision and adaptability in complex environments.

\section{RELATED WORKS}
\subsection{Legged Locomotion}  
Traditionally, legged locomotion relied on control-based methods\cite{gaertner2021collisions, chiu2022collision} to achieve basic locomotion tasks, and even gymnastic maneuvers\cite{li2024cafe}. 
Despite their effectiveness under controlled conditions, these methods often require precise modeling and manual tuning. 
With the rise of deep neural networks, learning-based approaches are now more robust and adaptable, allowing robots to perform a wider range of actions with minimal intervention. 
These methods have demonstrated significant advancements in robust locomotion\cite{margolis2024rapid, ji2022concurrent, fu2021minimizing}, agile motor skills\cite{hwangbo2019learning, he2024agile}, dynamic jumping\cite{margolis2021learning, li2023robust}, fall recovery\cite{yang2020multi, ma2023learning}, and complex parkour maneuvers\cite{zhuang2023robot}, as well as excelling in challenging terrains\cite{zhuang2023robot, cheng2024extreme} and confined spaces\cite{liao2023walking, miki2022learning}. 
While these advancements in locomotion are impressive, incorporating manipulation capabilities could substantially amplify the versatility and practical applications. 
\subsection{Quadrupedal Loco-manipulation} 
Recent advances in locomotion and manipulation have driven the development of integrated loco-manipulation systems, resulting in two main areas:


\textbf{Modular methods.} 
Several works proposed to divide loco-manipulation systems into distinct locomotion and manipulation components. These strategies often leverage off-the-shelf controllers\cite{unitree},\cite{bostondynamics} to manage mobility, while tailoring manipulation techniques for specific tasks\cite{zhang2024gamma, yao2022transferable, yao2023adaptive}. For instance, Yokoyama et al.\cite{yokoyama2023asc} shows adaptive skill coordination in diverse environments with robustness to dynamic obstacles and perturbations. 

Modular strategies simplify design by independently optimizing the arm for manipulation and the legs for locomotion, enhancing stability and flexibility.
However, treating the arm and body as distinct components 
can constrain the system's potential for achieving a more extensive workspace and integrated whole-body control.

\textbf{Unified methods.}
Meanwhile, other lines of work focused on achieving seamless loco-manipulation through unified whole-body control approaches. 
Zipeng et al.\cite{fu2023deep} initially proposed a unified policy for the simultaneous control of leg joints and manipulators. Building on this, Tifanny et al.\cite{portela2024learning} advanced the field with a whole-body force and position control. 
However, this approach inherits limitations from the earlier work\cite{fu2023deep}, including restricted arm orientation and reduced tracking accuracy. 
A two-stage policy was introduced to address these issues, but accuracy remains insufficient for complex tasks\cite{pan2024roboduet}.
Another SLAM-based method introduced a manipulation-focused whole-body controller, but it requires task-specific data and external processing, showing limited loco-manipulation capabilities\cite{ha2024umi}.
%
All these methods rely on teleoperation, requiring human intervention and limiting autonomy. Liu et al.\cite{liu2024visual} integrated high-level task planning with low-level control, achieving some autonomy.
However, this approach still struggles with low one-time grasping accuracy, limited generalization across tasks, and the need for extensive training.
%
In our work, we aim to overcome the existing issues by taking advantage of both modular and unified methods to develop a robust controller.


\section{METHOD}

In this section, we present our framework Generalizable Quadrupedal Whole-Body Grasping (QuadWBG). This framework is divided into four key components: locomotion, perception, manipulation, and planning (Figure \ref{fig:pipeline} A/B/C/D). 
First, a RL policy is trained to track 5D commands, allowing robust mobility. Meanwhile, the perception module generates real-time object masks and grasp poses, guiding the manipulation module as the arm transitions between tracking and grasping within the body frame. The planning module is trained based on proposed the Generalized Oriented Reachability Map (GORM) to optimize base positioning and enhance grasping performance. All policies are trained using Proximal Policy Optimization (PPO) in the Isaac Gym simulation.

The system is built on the Unitree B1 quadruped, equipped with a Unitree Z1 arm, Robotiq gripper, and an Intel RealSense D415 camera mounted on the wrist (Figure \ref{fig:pipeline}-E).

\subsection{Locomotion Module}\label{sec:locomotion_module}

An agile locomotion policy is essential for achieving high-accuracy in our whole-body loco-manipulation system. To implement this, we adopt the teacher-student architecture \cite{Lee_Hwangbo_Wellhausen_Koltun_Hutter_2020}, expanding the command set to five dimensions, including pitch ($\theta$) and height ($h$). Furthermore, to ensure stability and prevent undesirable behaviors, we constrain the sampling of command ranges based on height and pitch: $\theta \sim f(h), \; v_x, v_y, \omega \sim g(h, \theta)$.


\begin{table}[t!]
\centering\small
\begin{tabularx}{0.42\textwidth}{Xccc}  
\toprule
            Term                &  Min    &  Max &       Unit   \\
\midrule
\textbf{Dynamics}                           &            &        &  \;\;       \\
\hspace{3mm}Ground Friction                 & 0.1        & 3.00   &  \;\;   -    \\
\hspace{3mm}Base Payload Mass               & 0.0        &  20.0  &  \;  kg      \\
\hspace{3mm}Body Center of Mass$_{x}$       & -0.20      & 0.20   &  \;   m     \\
\hspace{3mm}Body Center of Mass$_{y,z}$     & -0.15      & 0.15   &  \;   m     \\
\hspace{3mm}Gripper Payload Mass            & 0.0        & 2.0    &  \;   kg     \\
\hspace{3mm}Motor Strength                  &  70        &  130   &  \;  \(\%\)  \\ 
\hspace{3mm}Proprioception Latency          &  0.005     &  0.045 &  \; s  \\    
\textbf{Rough Terrain}                      &            &        &  \;\;       \\
\hspace{3mm}Height                          &  0.0       &  0.1  &  \; m  \\   
\textbf{Observation}                        &            &        &  \;\;       \\
\hspace{3mm}Projected gravity noise         & -0.05      & 0.05   &  \;   -    \\
\hspace{3mm}Joint position noise            & -0.01      & 0.01   &  \;   rad     \\
\hspace{3mm}Joint velocity noise            & -1.5       & 1.5    &  \;   rad/s     \\
\hspace{3mm}Angular velocity noise          & -0.20      & 0.20   &  \;   rad/s    \\
\bottomrule
\end{tabularx}
\caption{Domain randomization settings for locomotion.}
\vspace{-2mm}  
\label{tab:low_level_dom_random}
\end{table}

The low-level teacher observation consists of proprioceptive and privileged observations, $o_{t}^{teacher} = (o_{t}^{prop}, o_{t}^{priv})$, where the proprioceptive observation  $o_{t}^{prop} \in \mathbb{R}^{64}$ includes the commands $c_{t}^{cmd} \in \mathbb{R}^{5}$, projected gravity vector $g_{t}^{base} \in \mathbb{R}^{3}$, base angular velocity $\omega_{t}^{base} \in \mathbb{R}^{3}$, joint positions and velocities $q_t,\dot{q}_t \in \mathbb{R}^{18}$, previous actions $a_{t-1} \in \mathbb{R}^{12}$, and phase variables $\theta_{t}^{feet} \in \mathbb{R}^{5}$ for natural walking patterns[wtw]. In contrast, the privileged observation $o_{t}^{priv} \in \mathbb{R}^{21}$ includes of parameters not accessible in real-time, such as base linear velocities, friction coefficient, mass parameters, and motor strengths.


We defined our reward functions to track commands and achieve efficient, natural walking. We penalize vertical body velocity and angular velocities around roll and pitch. The behavior rewards are primarily based on \cite{margolis2023walk}, \cite{ji2022concurrent}, we use the Raibert heuristic for foot placement, feet clearance to prevent tripping, and action smoothness for natural locomotion. Gait-conditioned and energy rewards further encourage stable, smooth gaits \cite{fu2021minimizing}. 

Locomotion policy is required to accurately track 5D commands under the perturbations caused by arm tracking and grasping the target objects. In order to facilitate this, we adopt height, and pitch invariant spherical coordinate-based sampling from \cite{fu2023deep}. Furthermore, we applied extensive domain randomization techniques
and observation noise to mitigate the sim-to-real gap in the robot’s dynamics Table \ref{tab:low_level_dom_random}. Among these strategies, adding proprioceptive latency proved crucial in preventing unstable motions caused by onboard computational delays. 


After training a robust teacher policy, we distill it into a student policy. The teacher uses a privileged information encoder to generate a latent vector $\bar{l}_t$, followed by a MLP to produce action $\bar{a}_t$. The student replaces the encoder with a GRU, which takes the proprioceptive observation $o_{t}^{prop}$ and generating a latent vector $l_t$, which is concatenated with $o_{t}^{prop}$ and passed through a MLP to produce action $a_t$. We use DAgger \cite{Ross_Gordon_Bagnell_2010} to supervise both $l_t$ and $a_t$., following \cite{Lee_Hwangbo_Wellhausen_Koltun_Hutter_2020}. The loss function is defined as:
\begin{equation}
L = (\bar{l}_t - l_t)^2 + (\bar{a}_t - a_t)^2
\end{equation}

\subsection{Perception Module} 

To achieve real-time tracking and precise grasp pose prediction, we utilized the Track Anything Model \cite{yang2023track} and ASGrasp \cite{shi2024asgrasp}. After generating the initial mask using SAM \cite{kirillov2023segment}, real-time object masks are provided by XMem \cite{cheng2022xmem}. ASGrasp takes inputs infrared (IR) images and RGB and can predict accurate depth, even for transparent and specular surface. 
The predicted depth point cloud is then input into GSNet \cite{wang2021graspness}, generating more accurate 6-DoF grasp poses. 

\subsection{Manipulation Module} 
The proposed manipulation module is designed to actively track and grasp target objects while simultaneously adapting the robot's base motion. We use a motion planning approach to address control inaccuracies in end-effector control with whole-body RL policies \cite{fu2023deep, portela2024learning}. The system operates in two distinct phases: tracking and grasping.

During tracking, we constrain the motion of the mounted camera within a predefined tracking sphere to minimize motion blur and prevent loss of tracking. As illustrated in \ref{fig:pipeline}-C, the tracking sphere is parameterized by its position and radius in the body frame. We use the Reachability Map (RM) to define the tracking sphere, ensuring that the camera operates exclusively within the dexterous workspace—where a valid inverse kinematics (IK) solution is available for arbitrary orientations.\cite{vahrenkamp2013robot, porges2015reachability}. For the IK solver, we apply a differential kinematics approach, as outlined in \cite{chi2023diffusion}, to ensure continuous and smooth motion. This technique minimizes redundant movements while maximizing vision coverage and overall motion smoothness.

The switching mechanism is built using the Reachability Map (RM) with a threshold-based reachability criterion. At each planning step, we calculate the reachability for the selected grasp pose using the RM. Once the threshold is met, the system switch to grasping phase. Our motion planner generates trajectories online, enabling the system to adapt to small unexpected movements while moving towards the target.




\begin{figure}[t]
\centering
\begin{overpic}
[bb=0 0 1366 956, width=0.74\linewidth]{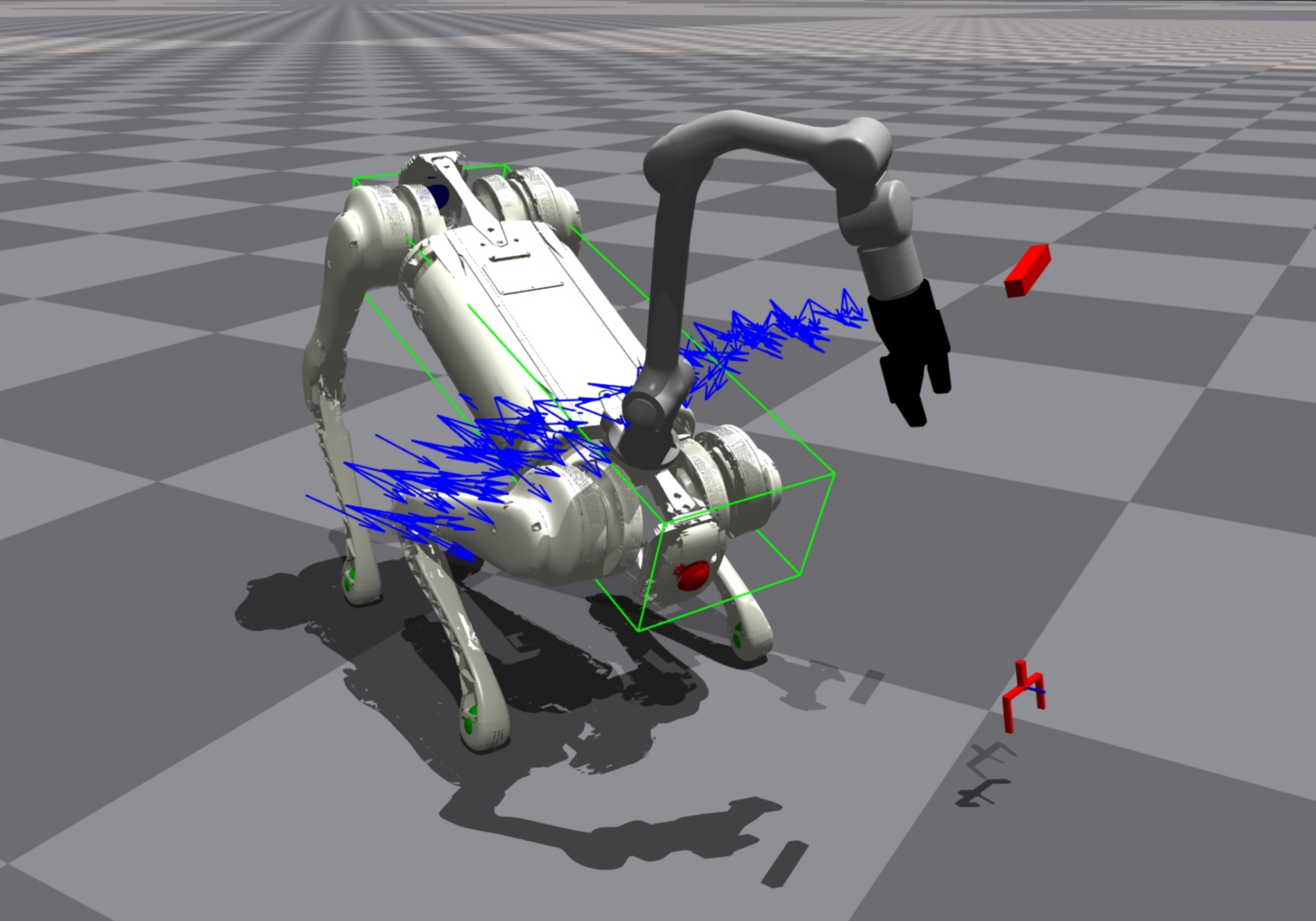}
\end{overpic}
\caption{ Visualization of GORM in simulation. Red gripper represent the target pose, blue arrows represent GORM, green cubic indicates the nearest base pose candidate.}
\vspace{-2mm}
\label{fig:GORM}
\end{figure}

\subsection{Planning Module}\label{sec:planning_module}
The Reachability Map (RM) is a pose quality metric commonly used to provide priors and objectives for mobile manipulation tasks \cite{jauhri2022robot, makhal2018reuleaux, jauhri2024active}. The Oriented Reachability Map (ORM) efficiently represents potential base poses relative to the Tool Center Point (TCP) frame \cite{vahrenkamp2013robot}. However, ORM is typically applied in platform-based mobile manipulation, where the robot's base is restricted to flat surfaces. We propose the Generalized Oriented Reachability Map (GORM), which supports robot base placements with six degree of freedom, as illustrated in Figure \ref{fig:GORM}. For any target pose in the world frame $p \in \text{SE}(3)$, the potential base-to-world distribution is computed via inverse of RM. We refine the space by removing base poses: 1) are out of locomotion range, 2) collide with the environment, and 3) fall below reachability threshold. Once a target pose is defined, GORM provides distribution of high quality potential base poses. We train the high-level policy to minimize the distance between current base pose to the nearest feasible pose:
\[r_\text{GORM} = \exp\left(-(\min \left(\mathbf{d}_\text{GORM}\right)\right)^2)\]

At every high-level step, we compute the distance distribution $\mathbf{d}_\text{GORM}$ by sum of Euclidean distance and geodesic distance between GORM and current base pose. We select the minimal distance from the distribution to encourage robot move close to the nearest base pose candidate. Since GORM is defined in the target pose frame, it only needs to be calculated once, making it highly efficient and well-suited for parallel training.



Training depth-based policies for obstacle avoidance \cite{zhuang2023robot, hoeller2021learning} and object grasping \cite{liu2024visual} is essential in vision-based control. However, these methods often introduce adding-noise and hole-filling techniques to compensate sim-to-real gap, which can degrade performance in precision tasks like grasping. Instead of relying on simulated depth images, our approach directly uses grasp pose detection results from an off-the-shelf perception module, as described in \cite{zhang2024gamma}. The high-level policy's observations are defined as:
\[
o_t = \left[ p_t^{(target)}, g_t^{(base)}, q_t, a_{t-1} \right].
\]
The vector \( p_t^{(target)} \in \mathbb{R}^{12} \) contains a rotation matrix \( \text{rot}_t \in \mathbb{R}^{9} \) and a translation vector \( \text{trans}_t \in \mathbb{R}^{3} \). The gravity vector in the body frame is \( g_t^{(base)} \in \mathbb{R}^{3} \), the joint positions are \( q_t \in \mathbb{R}^{18} \), and the last action is \( a_{t-1} \in \mathbb{R}^{5} \). The action comprises 5D commands \( c_{t}^{cmd} \in \mathbb{R}^{5} \), as discussed in Section \ref{sec:locomotion_module}.

\section{EXPERIMENTAL RESULTS}
We first conduct experiments in simulation and then real world to validate the performance of our system.
For our experiments, we deploy our system to on-board mini-computer with Intel Core i9-13900H and NVIDIA GeForce RTX 4070 Mobile (Figure \ref{fig:pipeline}-E).


\subsection{Evaluation in simulation} 
\textbf{Locomotion module evaluation.} 
We compared the performance and energy efficiency of several locomotion controllers, including AsymAC \cite{pinto2017asymmetric}, ROA \cite{fu2023deep}, StateEstimator \cite{margolis2023walk}, and Teacher-student \cite{hoeller2024anymal}. All methods were trained to track velocity commands under consistent conditions and were tested in a simulation environment with substantial randomization and observation noise to replicate real-world scenarios. Performance metrics were assessed using command tracking rewards and average torque, respectively. Given our focus on balancing low energy consumption and high performance, the teacher-student method offered an optimal trade-off and was selected as the core of our locomotion controller.

\begin{figure}[t!]
    \centering
    \begin{subfigure}[b]{0.21\textwidth}
        \centering
        \includegraphics[bb=0 0 1263 549, width=\textwidth]{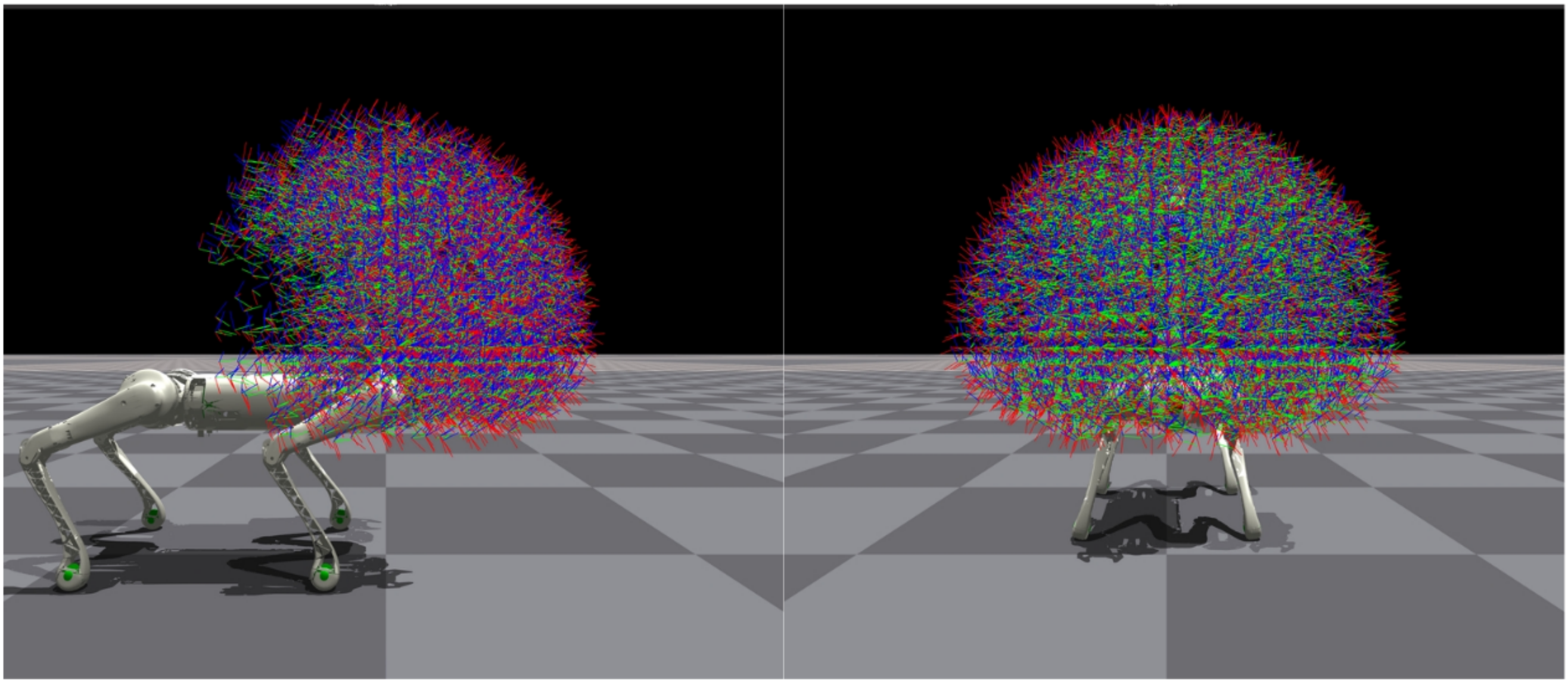}
        \caption{Arm workspace}
    \end{subfigure}
    \hspace{0.03mm}
    \begin{subfigure}[b]{0.21\textwidth}
        \centering
        \includegraphics[bb=0 0 1263 549, width=\textwidth]{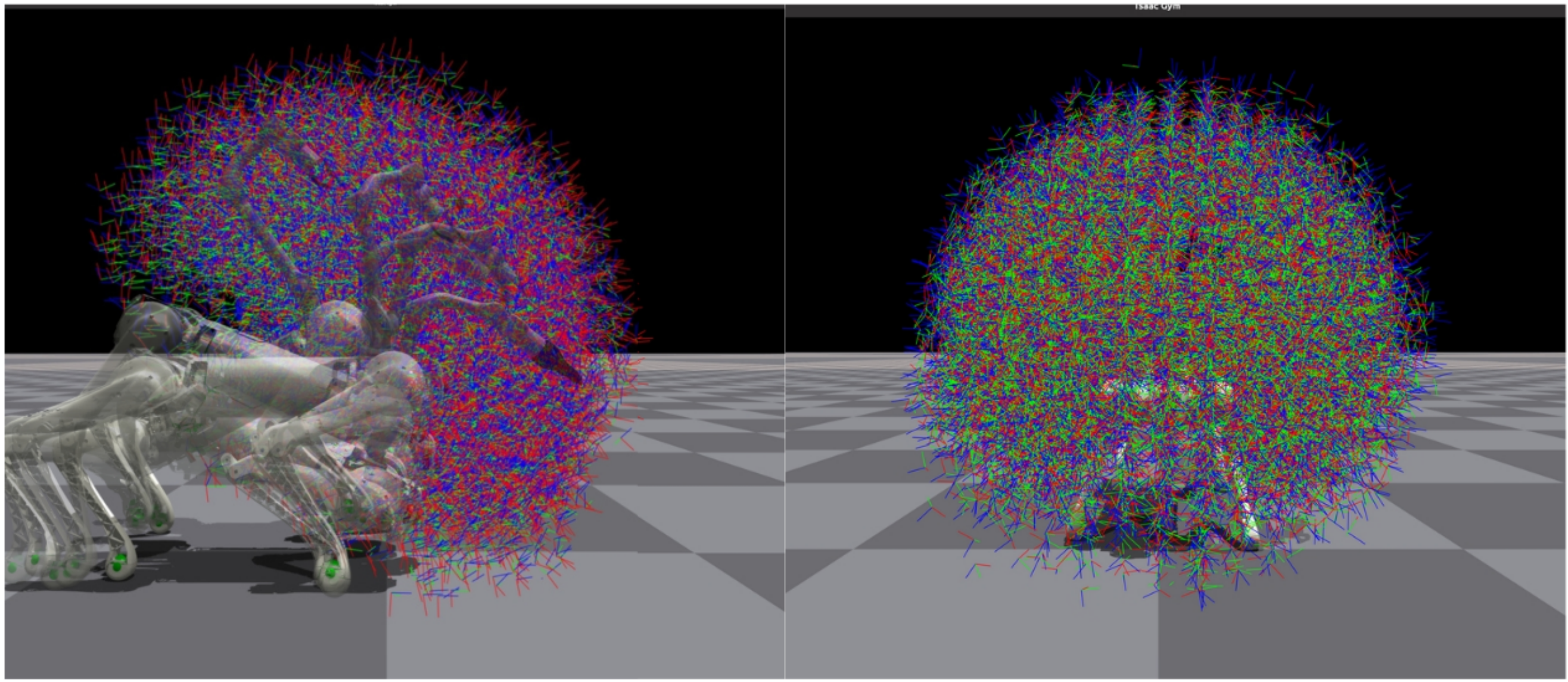}
        \caption{Whole body workspace}
    \end{subfigure}

    \vspace{3mm}

    \scriptsize
    \def\arraystretch{1}
    \centering
    \begin{tabularx}{\linewidth}{XXX}
        \hline
        & Workspace Volume [m\textsuperscript{3}] & Workspace Area [m\textsuperscript{2}] \\
        \hline
        Arm & 1.4 & 6.4 \\
        Whole body & 2.15 & 8.54 \\
        \hline
    \end{tabularx}

    \caption{The workspace of the fixed arm and the expanded workspace enabled by our whole-body control policy.}
    \label{fig:combined}
\end{figure}

\textbf{Whole-body workspace.} 
By tracking vertical commands, our locomotion policy, informed by arm proprioceptive feedback, extends the whole-body workspace without sacrificing robustness as illustrated in Figure \ref{fig:combined}. We measured the maximum reachable grasping target positions relative to the base, analyzing the convex hull properties for both arm and whole-body workspace. 
Our controller increased the efficient workspace volume by 54\%, and workspace area by 33\%, effectively following all commands across various body poses(forward and backward leans, crouching).

\begin{table*}[h!]
\centering
\small 
\begin{tabularx}{\linewidth}{c|c|XXXXXXX}
\toprule
& std & \multicolumn{7}{c}{Success Rate (\%) $\uparrow$} \\
& & Ball & Long Box & Square Box & Bottle & Cup & Bowl & Drill \\
\midrule
VBC & 14.53 & 55.40 & 28.57 & 80.00 & 56.57 & 68.01 & 58.96 & 53.33 \\
Ours w/o GORM & \textbf{2.16} & 57.09 & 55.85 & 56.47 & 50.66 & 55.85 & 52.52 & 54.81 \\
Ours & 3.46 & \textbf{91.66} & \textbf{89.67} & \textbf{90.67} & \textbf{81.33} & \textbf{89.67} & \textbf{84.33} & \textbf{88.00} \\
\bottomrule
\end{tabularx}
\caption{Success rates of our method compared to the baseline across seven object categories, tested in simulation. Each object underwent 300 trials.}
\label{tab:simluation_benchmark}
\end{table*}

\textbf{Generalizable whole-body grasping.}
We follow the same simulation benchmark setting as VBC\cite{liu2024visual} which use 34 objects divided into 7 categories: \textit{Ball, Long Box, Square Box, Bottle, Cup, Bowl, Drill}. In pick up experiment, we randomly reset the position and orientation of both robot and object at each round begin. We evaluate the success rate on each object for 300 trials, and the success condition is when the target object has been picked up before 150 high-level steps. We compare our method with no GORM reward and the baseline. For no GORM reward baseline, we applied the approach and assistant rewards from VBC. The aim of this comparison is to verify that our method can accurately perform mobile pick-up tasks across various object categories. As shown in Table \ref{tab:simluation_benchmark}, our method achieves a significantly higher success rate across all tested objects. Notably, the standard deviation for end-to-end RL methods like VBC is 14.53, reflecting considerable variability. VBC performs well on simple and small objects, such as the \textit{Square Box} (80\%), but struggles with more complex or larger objects, such as the \textit{Drill} (53.33\%) and \textit{Long Box} (28.57\%). In contrast, our modular method has a much smaller standard deviation of 3.46, indicating more consistent performance across all object types. Our approach demonstrates consistent performance regardless of the object's size or geometric complexity. This robustness can be largely attributed to the integration of the grasp pose detector, which provides less redundant input to the high-level policy. It allows the system to adapt more effectively to different objects, ensuring consistent performance even in challenging scenarios.
When the GORM reward is removed, we observe a significant drop in the success rate. This decline may be due to the basic reward approach being insufficient to lead the attached arm within the optimal workspace, especially in tasks that require precise manipulation skills.

\begin{table}[h!]
\centering
\small 
\setlength{\tabcolsep}{4pt} 
\begin{tabularx}{\linewidth}{c|>{\centering\arraybackslash}X>{\centering\arraybackslash}X|>{\centering\arraybackslash}X>{\centering\arraybackslash}X}
\toprule
 & VBC Objects on Floor & VBC Objects on Table & Transparent Objects & Arbitrary Objects \\
\midrule
GAMMA  & -- & 73\% & -- & -- \\
VBC  & 30\% & 12\% & -- & -- \\
Ours & \textbf{90\%} & \textbf{75\%} & \textbf{80\%} & \textbf{89\%} \\
\bottomrule
\end{tabularx}

\caption{Success rates of our one-time grasping method compared to the baseline across various object types and heights in real-world experiments. The floor and table heights are 0 meters and 0.5 meters.}
\label{tab:real_world_grasp_with_vbc}
\end{table}





\begin{figure}[t]
    \centering
    \includegraphics[bb=0 0 3543 1500, trim=100 100 50 10, clip, width=0.18\textwidth, angle=90]{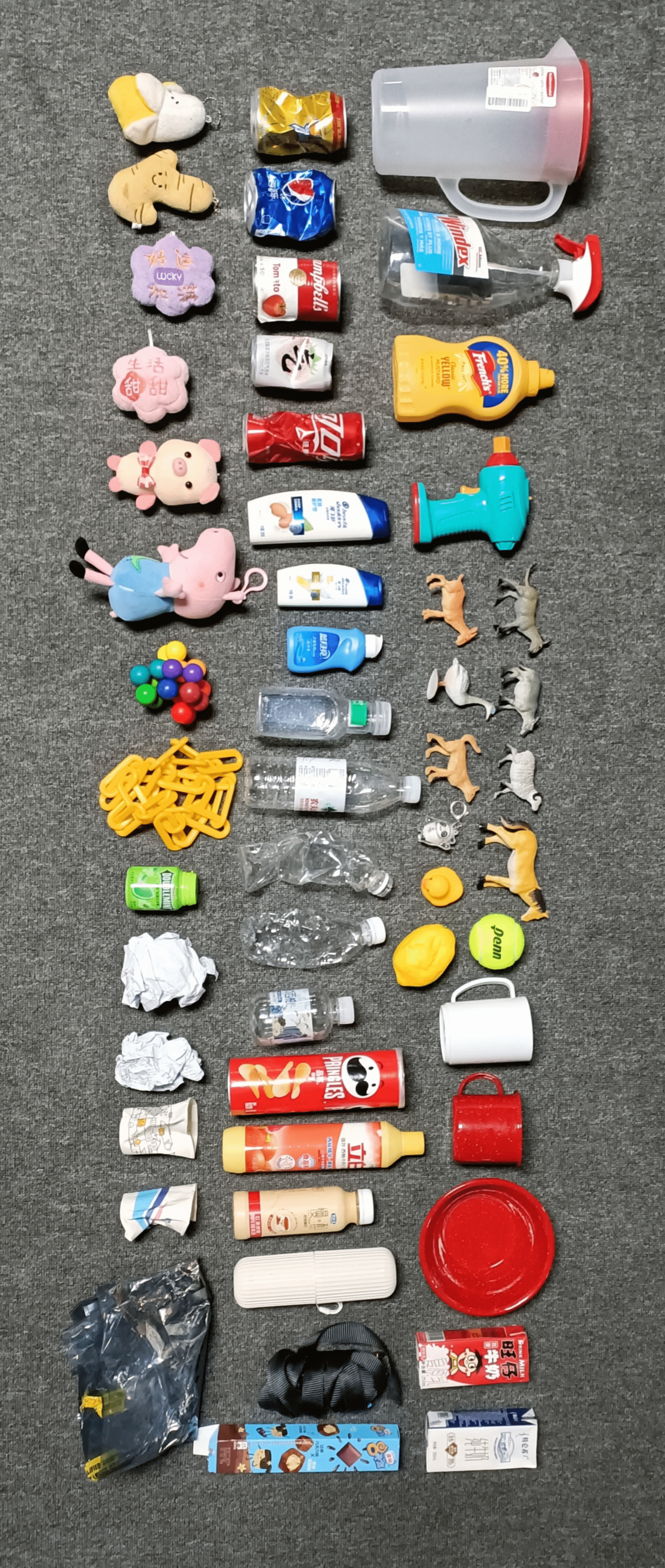}
    \caption{The illustration of the arbitrary household dataset used in our real-world tests, comprising items from the YCB dataset \cite{calli2017yale} as well as recyclables and toys. }
    \label{fig:dataset}
\end{figure}

\textbf{GORM guided grasping.} 
We use the same environment settings as described above, we use only \textit{Cup} as target object. We set four different height for target object each test for 500 trials: [0,0.3,0.75,1] meters. For each test, we perform 500 trials. After the 150th high-level step of each episode, we calculate the average reachability of the target pose relative to the body frame. We compare three policies: masked depth based policy, our policy without GORM reward, and our policy with GORM reward. Our policies take the observation mentioned in Section \ref{sec:planning_module}. For masked depth based policy, we replace the goal information \( p_t^{(target)} \) by the masked depth image which is obtained from the camera sensor on wrist in IsaacGym. The masked depth based and our method without GORM reward are trained based on approach and assistant rewards designed in VBC \cite{liu2024visual}. Shown in Table \ref{tab:high_level_reachability}, the masked depth-based policy consistently under-performs across all height levels, particularly struggling when the target is positioned lower than the base’s default range of motion. Without GORM, the policy shows a decline in reachability at different heights, as it fails to find the optimal base pose for grasping. In contrast, with GORM, the policy maintains high reachability across all heights by guiding the locomotion module to the best base position for grasping, maximizing performance regardless of object height. As shown in Table \ref{tab:simluation_benchmark}, incorporating GORM significantly improves the overall grasp success rate. This suggests that guiding the robot to an optimal base pose enables the manipulation module to perform more precise motion planning, resulting in a more robust and reliable system for grasping tasks.



\begin{table}[t!]
\centering
\small 
\begin{tabularx}{\linewidth}{c|XXXX}
\toprule
& \multicolumn{4}{c}{Reachability (\%) $\uparrow$} \\
& Floor & Box & Table & Shelf \\
\midrule
masked depth & 32.7 & 15.5 & 37.4 & 52.7 \\
w/o GORM & 63.3 & 55.5 & 45.6 & 51.4 \\
Ours  & \textbf{91.6} & \textbf{99.6} & \textbf{79.4} & \textbf{76.0} \\
\bottomrule
\end{tabularx}
\caption{Ablation study on Reachability at various heights: Floor(0.0 m), Box(0.3 m), Table(0.75 m), Shelf(1.0 m).}
\label{tab:high_level_reachability}
\end{table}


\begin{table}[t]
    \centering
    \setlength{\tabcolsep}{4pt} 
    \begin{tabularx}{\linewidth}{>{\centering\arraybackslash}X >{\centering\arraybackslash}X >{\centering\arraybackslash}X >{\centering\arraybackslash}X}
        \toprule
        Success Rate & Track Error & Grasp Error & Post-grasp Drop \\
        \midrule
        89.0\% & 6\% & 3\% & 2\% \\
        \bottomrule
    \end{tabularx}
    \caption{The results of our real-world trials, including the success rates and clear insights on performance issues.}
    \label{tab:sr_100_trials}
\end{table}

\subsection{Evaluation in real-world} 

\textbf{Comparison on VBC objects.}
We compare our system with end-to-end \cite{liu2024visual}, and modular method \cite{zhang2024gamma}. Former work has the ability to grasp objects on different heights, whereas the latter effectively deals with only table heights settings. 
We replicate 14 real world objects following \cite{liu2024visual}, 
randomly putting the object in random pose on floor and table. 
As shown in Table \ref{tab:real_world_grasp_with_vbc}, Our method shows the best performance across all tasks. While GAMMA\cite{zhang2024gamma} using model-based low-level controller, fails to grasp objects on the floor. VBC\cite{liu2024visual} shows a low success rate due to the RL based manipulation system.

\textbf{Experiment on arbitrary objects.}
Figure~\ref{fig:dataset} illustrates the dataset used in our real-world testing, which consists of objects sourced from both the YCB dataset \cite{calli2017yale} and commonly encountered recyclables, such as plastic bottles, cans  and crumpled paper or plastic packaging. Additionally, small toys frequently employed in grasping tests~\cite{wang2021graspness} are included in the dataset.  A significant portion of the dataset comprises transparent objects of various shapes.
During each trial, the dataset is shuffled, and a random subset of objects is dropped from a height of 1 meter to simulate random orientations and positions.
After the target object’s mask is simply annotated in the first frame, the automated grasping process is initiated. Table~\ref{tab:sr_100_trials} provides a summary of 100 trials conducted in the real-world setting. We observe our loco-manipulation system achieve 89\% success rate. This demonstrates the system’s generalizability across various objects, attributed to the modular design that combines an off-the-shelf perception model with robust, learned locomotion skills.

\textbf{Experiment on transparent objects.}
We conducted 10 trials on grasping transparent objects, as shown in Table \ref{tab:sr_100_trials}. Our results demonstrated 80\% grasping accuracy relying on single camera under cluttered scenario. This challenging task was tackled through the adaptive coordination of components, inheriting the ability from perception module without performance decay.

\section{CONCLUSION and LIMITATIONS}

In this paper, we introduced a modular whole-body manipulation system that integrates a learned 5D base locomotion policy with the novel Generalized Oriented Reachability Map (GORM) to achieve precise and robust loco-manipulation. Our approach effectively integrates accurate manipulation with coordinated whole-body locomotion, demonstrating significant improvements in handling a wide range of tasks in both simulation and real-world scenarios.

Despite recent advancements, the system still has several limitations. The planning module lacks collision avoidance and is not optimized for navigating challenging terrains. Additionally, it relies on human annotations rather than language inputs for target object grounding. These areas will be addressed in future work.
%

\clearpage


% ...










\bibliographystyle{IEEEtran}

\end{document}